\documentclass[9pt,technote,compsoc]{IEEEtran}
\ifCLASSOPTIONcompsoc
  \usepackage[nocompress]{cite}
\else
  \usepackage{cite}
\fi
\usepackage{graphicx}
\ifCLASSINFOpdf
\else
\fi
\usepackage{amsmath}
\usepackage{amssymb}

\usepackage{algorithmic}
\usepackage{algorithm}
\usepackage{subfig}
\usepackage{url}
\usepackage[pagebackref=false,breaklinks=true,letterpaper=true,colorlinks,bookmarks=false]{hyperref} 

\usepackage{times}
\usepackage{multirow}

\newcommand{\etal}{\emph{et al.}}
\newcommand{\ie}{\textit{i}.\textit{e}., }
\newcommand{\eg}{\textit{e}.\textit{g}., }
\def\mat#1{\mathchoice{\mbox{\boldmath $\displaystyle\tt#1$}}
{\mbox{\boldmath$\textstyle\tt#1$}}
{\mbox{\boldmath$\scriptstyle\tt#1$}}
{\mbox{\boldmath$\scriptscriptstyle\tt#1$}}}

\def\vect#1{\mathchoice{\mbox{\boldmath $\displaystyle\bf#1$}}
{\mbox{\boldmath  $\textstyle\bf#1$}}
{\mbox{\boldmath  $\scriptstyle\bf#1$}}
{\mbox{\boldmath  $\scriptscriptstyle\bf#1$}}}

\def\vh{{\vect h}}

\def\vp{{\vect p}}

\def\vv{{\vect v}}
\def\vw{{\vect w}}
\def\vx{{\vect x}}

\def\vX{{\vect X}}

\def\v0{{\vect 0}}

\def\mDelta{{\mat\mDelta}}

\def\m1{{\mat 1}}

\def\Reales{\mathbb{R}}

\IEEEpubid{
\fbox{
\parbox{0.975\textwidth}{
\copyright 2021 IEEE. Personal use of this material is permitted. Permission from IEEE must be obtained for all other uses, in any current or future media, including\hfill
reprinting/republishing this material for advertising or promotional purposes, creating new collective works, for resale or redistribution to servers or lists, or reuse of\hfill
any copyrighted component of this work in other works. \vspace{-0.5mm}
}%
}}

\begin{document}
%
\title{Multi-task head pose estimation in-the-wild}
%
%
%

\author{Roberto~Valle,~
        Jos\'e~M.~Buenaposada~
        and~Luis~Baumela
\IEEEcompsocitemizethanks{\IEEEcompsocthanksitem Roberto Valle and Luis Baumela are with the Departamento de Inteligencia Artificial, Universidad Polit\'ecnica de Madrid, Campus de Montegancedo s/n, 28660 Boadilla del Monte, Spain.\protect\\
E-mail: rvalle@fi.upm.es and lbaumela@fi.upm.es
\IEEEcompsocthanksitem Jos\'e. M. Buenaposada is with the ETSII, Universidad Rey Juan Carlos, C/ Tulip\'an s/n, 28933 M\'ostoles, Spain.\protect\\
E-mail: josemiguel.buenaposada@urjc.es
}
}

%
%

\markboth{Journal of \LaTeX\ Class Files,~Vol.~14, No.~8, August~2015}%
{Shell \MakeLowercase{\textit{et al.}}: Bare Demo of IEEEtran.cls for Computer Society Journals}
%



\IEEEtitleabstractindextext{%
\begin{abstract}
We present a deep learning-based multi-task approach for head pose estimation in images.
We contribute with a network architecture and training strategy that harness the strong dependencies among face pose, alignment and visibility, to produce a top performing model for all three tasks.
Our architecture is an encoder-decoder CNN with residual blocks and lateral skip connections. 
%
%
We show that the combination of head pose estimation and landmark-based face alignment significantly improve the performance of the former task. 
Further, the location of the pose task at the bottleneck layer, at the end of the encoder, and that of tasks depending on spatial information, such as visibility and alignment, in the final decoder layer, also contribute to increase the final performance.
%
In the experiments conducted the proposed model outperforms the state-of-the-art in the face pose and visibility tasks.
By including a final landmark regression step it also produces face alignment results on par with the state-of-the-art.
\end{abstract}

}

\maketitle

\IEEEdisplaynontitleabstractindextext

%
\IEEEpeerreviewmaketitle


\section{Introduction}\label{sec:introduction}

%
%
%
%

\IEEEPARstart{H}ead pose greatly affects facial appearance. It is one of the parameters that influences to a largest extent the performance of many face analysis tasks. For this reason it is a fundamental step in computer vision algorithms estimating attention~\cite{Bergasa08}, identifying social interaction~\cite{Ba11}, recognizing faces~\cite{Chang17} or robustly estimating facial attributes~\cite{Kumar18b,Ranjan19}.
It is a challenging problem in ``in-the-wild'' conditions, \ie\ in presence of extreme orientations, partial occlusions and varying resolution, illumination, facial hair and makeup. Although it has been often considered as by-product or auxiliary task of facial landmark location~\cite{Wu19}, recent results prove that it is much more efficient than landmark estimation and it may achieve superior performance in subsequent face analysis tasks, such as recognition~\cite{Chang17}.
In this paper we present a multi-task approach to head pose estimation in unrestricted images. We exploit the strong dependencies among head pose and landmark-related tasks within a multi-task Convolutional Neural Network (CNN) to produce a top performing model.

The \emph{multi-task learning} (MTL) paradigm encompasses a set of learning techniques that provide effective mechanisms for sharing information among multiple tasks. 
It enables the use of larger and more diverse data sets, that improve the regularization during training and the generalization of the final model~\cite{Caruana97}. 
MTL is intimately related to \emph{transfer learning} (TFL). In TFL a model is trained for one or more auxiliary tasks and subsequently refined for a main target task~\cite{Razavian14,Zamir18}. Traditionally MTL implies a simultaneous or parallel treatment of all tasks~\cite{Caruana97}, whereas in TFL tasks are learned sequentially. In our approach we combine parallel and sequential learning, so it cannot be clearly cast into one of the above two schemes. We rather generalize the traditional concept of MTL to include both. Following other approaches in the literature~\cite{Lee16} we consider different degrees of MTL asymmetry.  In this regard TFL is an extremely asymmetric MTL scheme in which auxiliary tasks are only used for pre-training. 
In our proposal we adopt an asymmetric approach where we seek to optimize head pose using visibility and alignment as auxiliary tasks. However, as we show in our experiments, the co-operation in our model among all three tasks is so high that all of them achieve state-of-the-art results in the most popular benchmarks and improve the performance they would otherwise achieve independently.

A key element in a multi-task CNN is the architecture of the model and the location of each task in the net. A natural approach is to share bottom layers among all tasks, since they model low-level features, whereas top layers, that capture high level features, are specific to each task~\cite{Kokkinos17}. In the context of face processing, some approaches have completely separate networks to model each attribute~\cite{Chang19}, others share all features in a common backbone~\cite{Han18}, and others combine feature maps from different parts of the encoder network~\cite{Ranjan19}. In our architecture, an encoder-decoder CNN, we carefully place each task to optimize the final performance. We locate head pose, a holistic task, at the end of the encoder. In this way the network bottleneck acts as embedding representing face pose. Visibility and alignment tasks are located at the decoder end, since they require information about the spatial location of landmarks in the image.

To train our model we leverage on the large face landmarks annotated data sets available. We first train the CNN for the landmarks-based face alignment task. Then we fine-tune it for head pose, face alignment and landmarks visibility. In the most asymmetric incarnation of our model, once trained, we may dispose of the decoder and the associated alignment and visibility tasks to produce a very efficient head pose estimation system. 
Alternatively, we may keep the full trained model and use the landmarks visibility and alignment outputs of the CNN as input to a novel face landmarks regression module based on an ensemble of regression trees. This model further improves the accuracy of landmarks location by imposing a valid face shape on the set of regressed landmarks.

We evaluate our model for all three tasks using COFW, AFLW and AFLW2000-3D landmark-based data sets. In these experiments our model beats the top competing approaches in the head pose and visibility estimation tasks. It also achieves performance comparable with the state-of-the art for the landmark-based face alignment task. We also evaluate head pose estimation with \emph{Biwi}. Although it was not acquired in-the-wild, this data set is a widely used marker-less benchmark for head pose estimation. Here our results also establish a new state-of-the-art. 

\begin{figure*}
  \centering
  \includegraphics[width=0.32\textwidth]{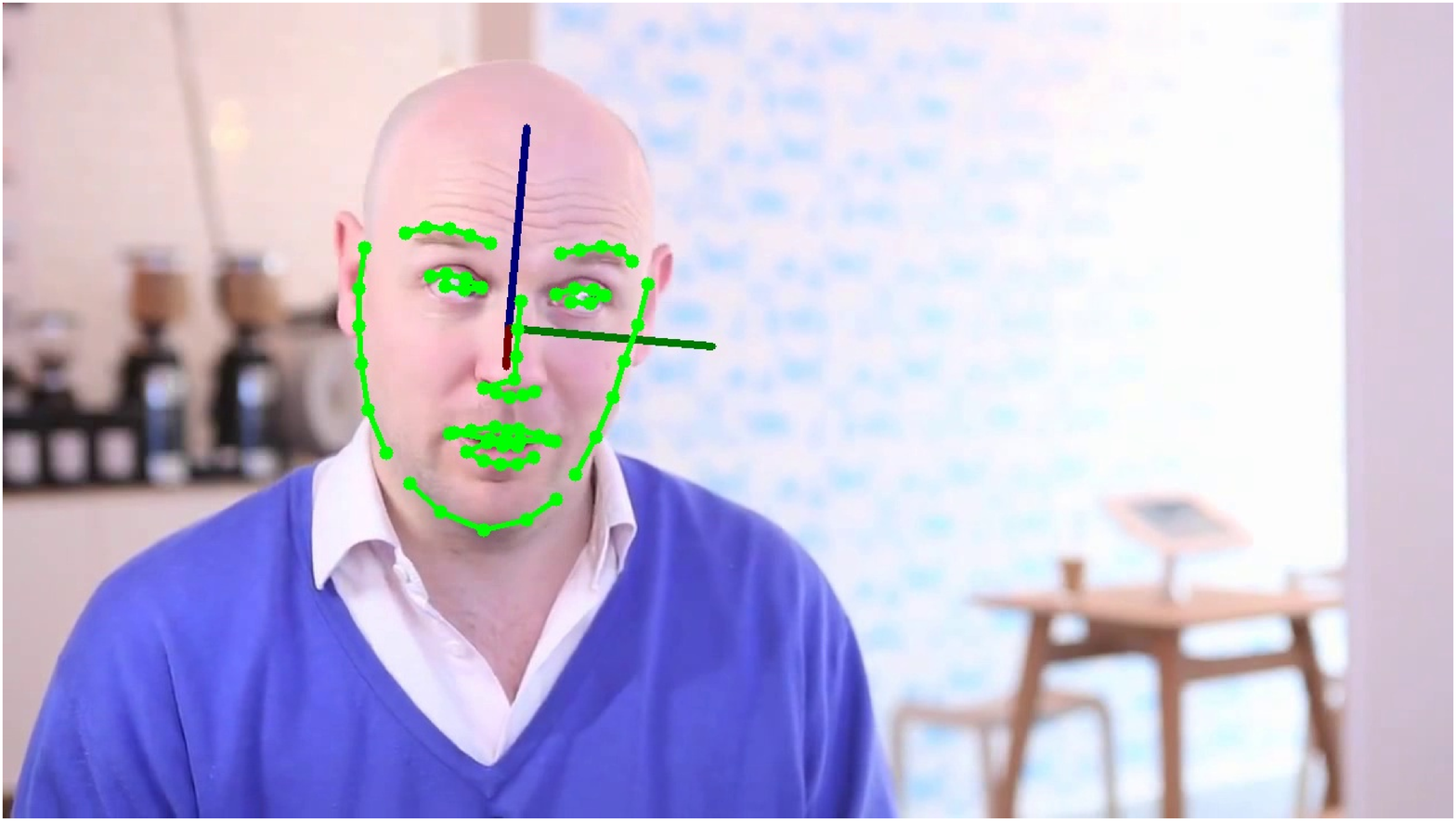}
  \includegraphics[width=0.32\textwidth]{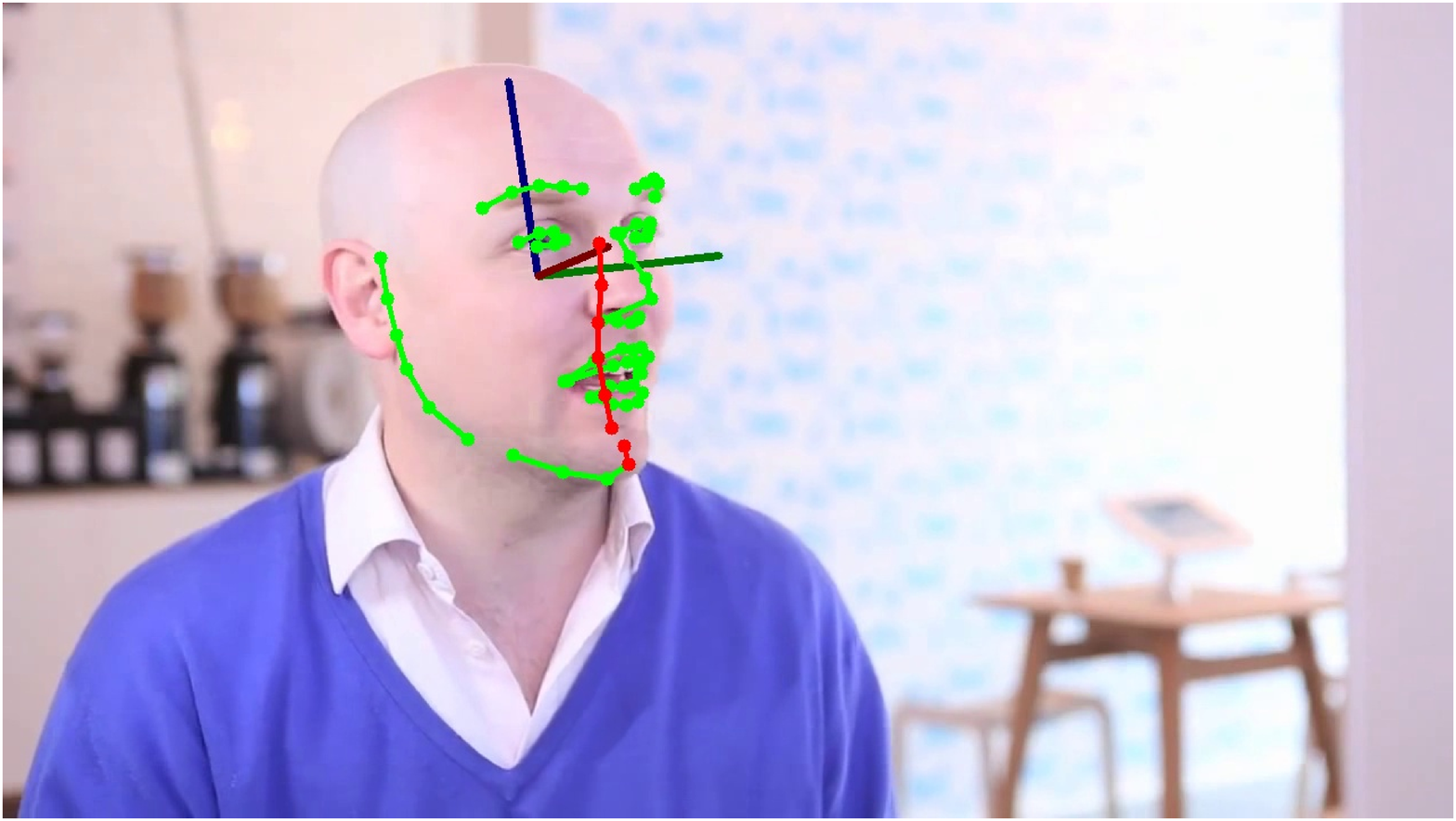}
  \includegraphics[width=0.32\textwidth]{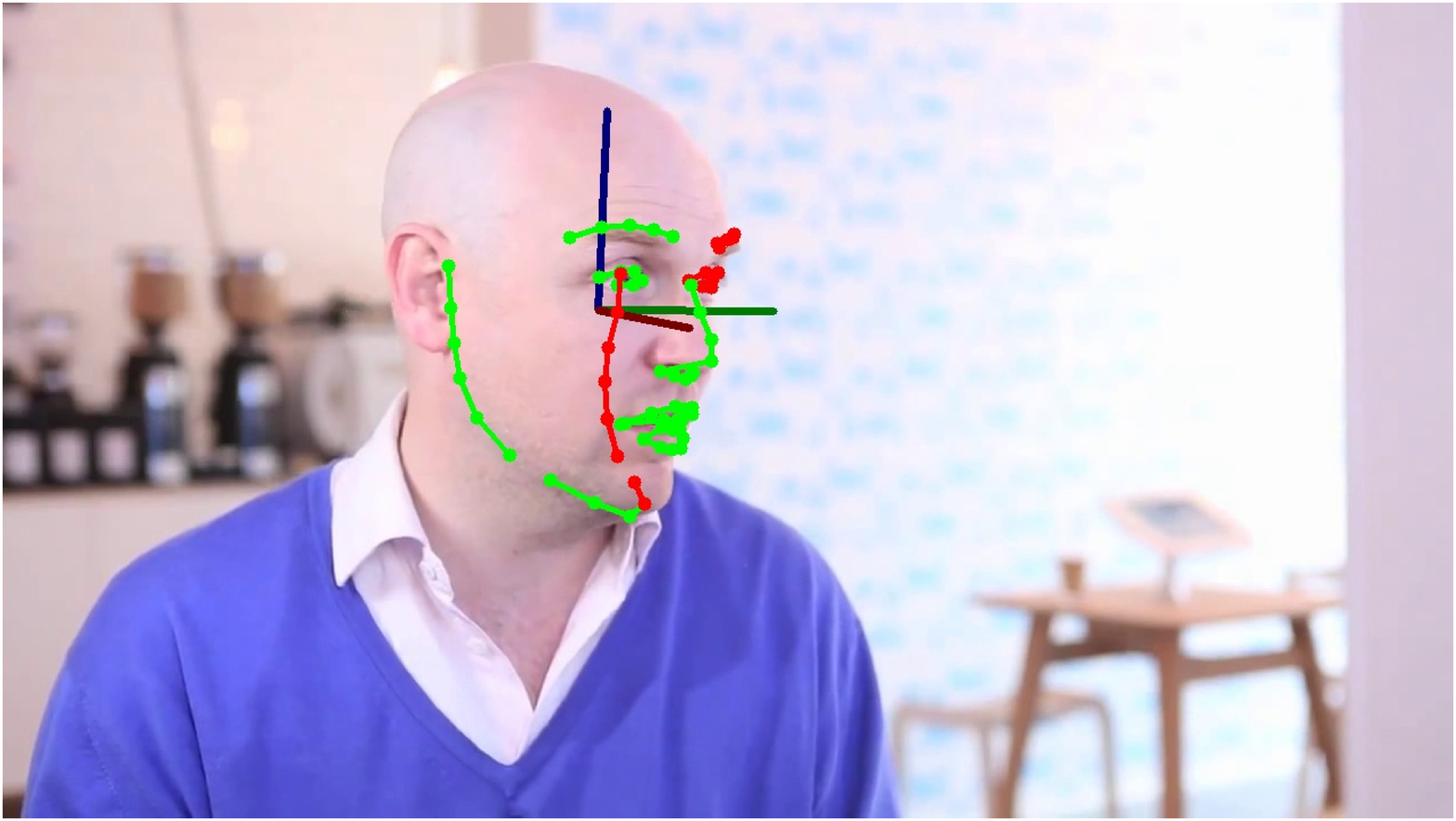}
  \caption{Simultaneous head pose estimation, facial landmark location and their visibility predictions when processing a video from 300VW~\cite{Shen15}. Green and red points show visible and non-visible landmarks respectively. The co-ordinate system qualitatively represents head pose.}
  \label{fig:multitask_video}
\end{figure*}

In summary, we propose a multi-task approach for head pose estimation. The proposed solution combines a good model architecture, training strategy and a set of complementary tasks that boost final performance. The resulting model achieves top results for all three tasks, head pose, face alignment and visibility.
In Fig.~\ref{fig:multitask_video} we display our predictions for some frames of a video from 300VW~\cite{Shen15}. It shows a remarkable tracker-like stability although each frame is processed independently.


\section{Related Work}
\label{sec:related_work}

The unique ability of neural networks to transfer and share knowledge among various tasks is one of the reasons for its present success. This is typically done using MTL techniques.
In computer vision MTL has been widely used to simultaneously learn related tasks such as semantic segmentation and surface normal prediction~\cite{Kokkinos17}.
%
In the facial analysis field, head pose is often used as a pre-processing step to help estimate 
face landmarks~\cite{Dantone12,Yang15,Kumar18a}.
%
Other approaches simultaneously estimate head pose with facial landmarks~\cite{Kumar18b,Ranjan19}, Facial Action Units~\cite{Zhou17}, gender~\cite{Ranjan19} and various other facial attributes~\cite{Ranjan17}.
%
Alternatively, facial attributes estimation have also been combined with landmark detection~\cite{Zhang16,Ranjan17,Han18}.
%
%
In our approach we follow an asymmetric MTL scheme where the primary task is head pose estimation and use face landmarks as an auxiliary task that regularize and improve the performance of the primary task. 


Pre-training a deep model with a large and general data set such as ImageNet has been a common practice for multiple vision tasks~\cite{Razavian14}. 
%
In the context of face analysis, ImageNet~\cite{Amador17,Ruiz18,Hsu18,Ranjan19,Liu19b,Zhang20}, and other large face-related data sets, such those for face recognition~\cite{Liu15, Zhong16,Ranjan17,Cao18b}, have also been extensively used for predicting various facial attributes.
More recently, self-supervised tasks have also emerged as powerful unsupervised pre-training mechanisms~\cite{Wiles18b,Zhang18e,Thewlis19,Jeon19}. 
For estimating head pose, pre-training with an unsupervised face alignment task yields better results than using a large supervised face recognition data set~\cite{Wiles18b}. This is possibly due to the geometrical cues learned in the alignment process. 
Following the same reasoning, we hypothesize that face landmark estimation is related to head pose. So, pre-training with the former task may improve the performance of the latter. Moreover, there is a lack of annotated ``in-the-wild'' head pose data sets. With our approach we leverage on the abundance of in-the-wild landmark-annotated data to train our model. As we show in the experiments, pre-training with a facial landmark estimation task improves head pose accuracy, beating other  ImageNet pre-trained competing models~\cite{Amador17,Ruiz18,Hsu18,Liu19b}.

In a MTL strategy the final results depend on the affinity or degree of co-operation among the tasks involved~\cite{Zamir18,Standley19}. In extreme situations \emph{negative transfer} may actually hinder the final performance~\cite{Wang19,Lee16}. 
Many approaches that simultaneously estimate head pose with other facial attributes, \eg~\cite{Kumar18b,Ranjan17,Ranjan19}, combine various competing tasks in the same network layer. In our experiments we show that head pose does not co-operate with landmark-related tasks when placed in the same layer.  To address this issue we propose to use  an  encoder-decoder  CNN  and  locate  head pose,  a holistic task, at the encoder end, that represents global face information. We place landmark-related tasks
at the decoder end, where spatial information is represented at the finest detail (see Fig.\ref{fig:multitask}). 

The best head pose estimation algorithms address the problem from a single task perspective.
In the simplest case they fine-tune a backbone previously trained on ImageNet~\cite{Amador17,Chang17,Ruiz18}. 
%
QuatNet and GLDL are respectively the state-of-the-art in AFLW and AFLW2000-3D. They use standard CNN-based models pre-trained in ImageNet. QuatNet combines ordinal and L2 regression losses representing head pose angles with quaternions~\cite{Hsu18}.
GLDL learns a Gaussian distribution per co-ordinate using a Gaussian Labels Distribution Loss (GLDL)~\cite{Liu19b}. 
FDN and FSA-Net are the top performers in the Biwi data set. Both approaches stand on specifically taylored network architectures. FDN uses a three-branch network with a feature decoupling module to explicitly learn discriminative features for each pose angle\cite{Zhang20}. FSA-Net combines spatially grouped pixel-level features of activation  maps from different layers~\cite{Yang19}. A recent alternative achieves state-of-the-art results on Biwi training with synthetically generated data~\cite{Kuhnke19}. To this end it introduces an adversarial domain adaptation approach for partially shared and continuous label spaces.

We leverage on the ideas discussed above to build a top performing head pose estimation algorithm. Our architecture is an standard encoder-decoder CNN with residual blocks and lateral skip connections. The key element of our proposal is a MTL scheme that combines a set of complementary tasks strategically located in the architecture. With our approach we improve not only the prediction accuracy, but also the computational and data efficiency, compared to training different models with data sets for each task.


\section{Multi-task head pose estimation}
\label{sec:algorithm}

In this section we present our two-stage framework termed MNN+OR. First, we describe a novel Multi-task CNN (MNN) that estimates head pose, landmark heatmaps and their visibilities (see Fig.~\ref{fig:multitask}). Second, we introduce an Occlusion-aware Regressor (OR) that we use to regress the location of facial landmarks (see Fig.~\ref{fig:oert}). 

\subsection{Multi-task Neural Network (MNN)}
\label{sec:mnn}

The most successful CNN architectures for facial landmark detection use an encoder-decoder network with lateral connections such as U-Net~\cite{Ronneberger15} and RCN~\cite{Honari16}. Both capture local and global features at different scales. The popular Hourglass architecture~\cite{Yang17} has a similar topology with extra convolutional layers in the lateral connections. 

\begin{figure}
\centering
\includegraphics[width=0.49\textwidth]{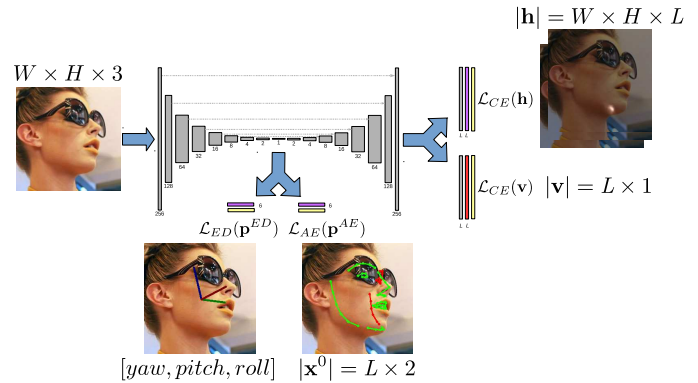}
\caption{Multi-task encoder-decoder for the estimation of head pose, $\mathcal{L}_{ED}(\vp^{ED})$, rigid and deformable facial landmarks location, $\mathcal{L}_{AE}(\vp^{AE})$ and  $\mathcal{L}_{CE}(\vh)$, and their visibilities, $\mathcal{L}_{CE}(\vv)$. We locate the head pose and rigid landmarks estimation tasks at the bottleneck layer, and the non-rigid face deformation and visibilities at the decoder end.}
\label{fig:multitask}
\end{figure}

In this section, we introduce an architecture termed Multi-task Neural Network (MNN) based on a U-Net encoder-decoder with bottleneck residual blocks~\cite{He16b} instead of its original convolutional layers. The residual block lets us reduce the number of operations and increase depth while preserving the gradient back propagation through. We also include lateral skip connections that link symmetric layers between the encoder and the decoder preserving the spatial information (see the Supplementary Material). 

MNN is a symmetric encoder-decoder architecture each with 9 stages. The encoder reduces the spatial extent of the input face image from $256\times 256$ to $1\times 1$ pixels. In the depth dimension we increase the number of feature maps from 64 in the first layer up to 256 in the bottleneck. 
We also include BatchNormalization and ReLu after each convolutional layer.

We encourage the encoder to act as feature embedding that learns a holistic face representation, favouring the exchange of information among all tasks. We attach to this layer two losses related to head pose estimation. The decoder learns local features tailored to the estimation of non-rigid landmark locations and their visibilities. Henceforth, we describe these losses and tasks.

\textbf{Holistic tasks.} 
The location of the loss functions associated to our tasks is essential given that the feature maps in different layers of the CNN represent the image information at different levels of abstraction and aggregation.

Since the head pose is a global attribute, we compute it from the bottleneck layer at the encoder end. Our objective here is to estimate the six parameters of the rigid transformation, $\vp\in\Reales^6$, representing the relative pose between the head and the camera. To this end, we include two fully connected layers, $\vp^{ED}$ and $\vp^{AE}$, with 6 outputs each at the end of the encoder (see Fig.~\ref{fig:multitask}). We optimize these layers with two loss functions,
\begin{equation}
   \mathcal{L}_{ED}(\vp^{ED}) = 
   \sum_{i=1}^{N}
   || \tilde{\vp}_i - \vp_i^{ED} ||_2,
\label{eq:euclidean_distance_loss_min}
\end{equation}
\begin{equation}
   \mathcal{L}_{AE}(\vp^{AE}) = \sum_{i=1}^{N}
   \left(
   \sum_{l=1}^{L}
      \left(
       \frac{\tilde{\vw}_i^l}{||\tilde{\vw}_i||_1} \cdot{} ||\tilde{\vx}_i^l - \pi(\vp_i^{AE},\vX^l)||_2
      \right)
   \right),
\label{eq:loss_l2_proj}
\end{equation}
where $N$ denotes the number of images, $L$ the number of landmarks, $\vp_i^{ED}$ and $\vp_i^{AE}$ the predicted pose parameters for the $i$-th training image using each loss, $\tilde{\vp}_i$ the ground truth head pose parameters for the $i$-th training image, $\tilde{\vx}_i^l\in\Reales^{L\times 2}$ the $l$-th landmark ground truth co-ordinates for the $i$-th training image, $\vX^l\in\Reales^{L\times 3}$ the 3D co-ordinates of the $l$-th landmark, and $\pi$ the camera projection.

Each loss plays an important role in our model. On the one hand, $\mathcal{L}_{ED}(\vp^{ED})$ directly minimizes the euclidean error of pose parameters and provides an accurate and unambiguous pose estimation, $\vp^{ED}$. On the other hand, $\mathcal{L}_{AE}(\vp^{AE})$ measures the alignment error produced by the rigid projection of the mean 3D face model, $\vx_i=\pi(\vp_i^{AE},\vX)$. The latter provides a better landmark initialization for the OR stage. However, the pose estimated, $\vp^{AE}$, has projection ambiguity and estimation error caused by $\vX$ not being the actual 3D landmark location, but that of the mean face. The combination of both losses provides unambiguous and accurate pose regression, as well as accurate rigid landmark localization. 

\textbf{Position-dependent tasks}. 
Facial landmarks detection and their visibility estimation require both global and abstract features with a fine spatial resolution. Therefore, we use the feature maps at the end of the MNN decoder to estimate these attributes (see Fig.~\ref{fig:multitask}). For the landmark location task we introduce a convolutional layer producing [$256 \times 256 \times L$] feature maps and a softmax activation layer to generate heatmaps, such that $\sum_p^{256\times 256}\vh(p)=1$. For the visibility task, we add a pooling layer with kernel size $256\times 256$ to generate the vector of $L$ visibilities associated to our landmarks, $\vv$. To train this model we use the cross-entropy loss,
\begin{equation}
   \mathcal{L}_{CE}(\vh) = \sum_{i=1}^{N}
   \left(
   \sum_{l=1}^{L}
      \left(
      \frac{\tilde{\vw}_i^l}{||\tilde{\vw}_i||_1} 
      \sum_{p=1}^{256\times256}
      \left(
      -\tilde{\vh}_i^l(p) \cdot{} \log(\vh_i^l(p))
      \right)
      \right)
   \right),
\label{eq:loss_crossentropy}
\end{equation}
\begin{equation}
   \mathcal{L}_{CE}(\vv) = \sum_{i=1}^{N}
   \left(
      \sum_{l=1}^{L}
      \left(
      \frac{\tilde{\vw}_i^l}{||\tilde{\vw}_i||_1} 
      \sum_{p=1}^{2}
      \left(
      -\tilde{\vv}_i^l(p) \cdot{} \log(\vv_i^l(p))
      \right)
      \right)
   \right),
\label{eq:loss_crossentropy_vis}
\end{equation}
where $N$ is the number of images, $L$ the number of landmarks, $\tilde{\vh}_i^l$, $\vh_i^l$ the $l$-th ground truth and predicted heatmaps for the $i$-th training image, and $\tilde{\vv}_i^l$, $\vv_i^l$ the $l$-th ground truth and predicted visibilities for the $i$-th training image.

To handle unlabelled landmarks we include $\tilde{\vw}^l$, a landmark mask indicator variable ($\tilde{\vw}_i^l=1$ when the $l$-th landmark is annotated, and $\tilde{\vw}_i^l=0$ otherwise). This loss also enables data augmentation with large rotations, translations and scales, labelling landmarks falling outside of the bounding box as missing ($\tilde{\vw}_i^l=0$).

\textbf{Multi-task loss.} 
The loss function $\mathcal{L}(\vp^{ED}, \vp^{AE}, \vh, \vv)$ computes a global error obtained from the pose parameters $\vp^{ED}$, $\vp^{AE}$, the landmark heatmaps, $\vh$, and the visibilities, $\vv$, by combining them using a weighted sum of the losses,
\begin{equation}
\begin{split}
   \mathcal{L}(\vp^{ED}, \vp^{AE}, \vh, \vv) =\;&\alpha_{\vp_{1}}\mathcal{L}_{ED}(\vp^{ED}) + \alpha_{\vp_{2}}\mathcal{L}_{AE}(\vp^{AE})\;+\\
   &\alpha_{\vh}\mathcal{L}_{CE}(\vh) + \alpha_{\vv}\mathcal{L}_{CE}(\vv).
\end{split}
\label{eq:global_loss}
\end{equation}

We empirically tune the weights $\alpha_{\vp_{1}}$, $\alpha_{\vp_{2}}$, $\alpha_{\vh}$ and $\alpha_{\vv}$ to balance the importance of all tasks. 
To this end, we train each task individually and determine the relative loss magnitudes when the learning process converges and ponder them accordingly.



\subsection{Occlusion-aware Regressor (OR)}
\label{sec:oert}

To achieve top results in the facial landmarks detection task we use an Ensemble of Regression Trees (ERT) that regularizes the MNN result by enforcing it to be a valid face shape~\cite{Cao14}. To this end, we introduce an Occlusion-aware Regressor (OR). It is different from other landmark regressors in the literature~\cite{Burgos13,Wu17,Valle18,Valle19b} in that our approach leverages on the robust landmark location and visibility estimation available at the MNN decoder to regress the landmark co-ordinates with top accuracy.

\begin{figure}
\centering
\includegraphics[width=0.49\textwidth]{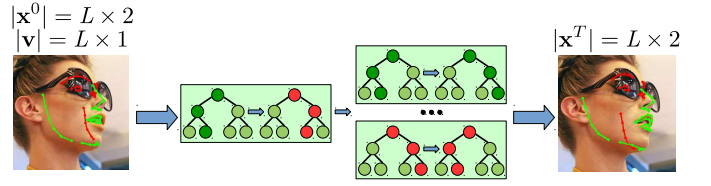}
\caption{The OR is initialized with the 3D face model projected landmarks, $\vx^0$, and their visibilities, $\vv$. It incrementally updates the landmark location discarding the predictions of those regression trees whose features are extracted around occluded landmarks, shown in red.}
\label{fig:oert}
\end{figure}

\textbf{OR initialization.} We use the head pose estimated by the MNN (see Section~\ref{sec:mnn}) to project AFLW mean 3D face model onto the image using $\vx^0_i=\pi(\vp_i^{AE},\vX)$, a $L\times 2$ matrix (see Fig.~\ref{fig:oert}). This provides the OR with an initial estimation of the scale, and position of the target face shape. With this initialization we ensure that $\vx^0_i$ is a valid face shape. This guarantees that the predictions in the next step of the algorithm, using an ERT, will also be valid face shapes~\cite{Cao14}. Here, we also initialize the visibilities according to the head pose (\ie self-occlusions due to extreme head pose orientations) and the MNN prediction (\ie occlusions), instead of regressing the visibility in the ERT cascade like~\cite{Burgos13,Valle18,Valle19b}. 



\textbf{Non-rigid face shape deformation.} Since the OR is initialized with the rigid face shape in the correct pose  (see Fig.~\ref{fig:oert}), to align the face it only needs to estimate the remaining non-rigid deformation of the face. To handle occlusions we incorporate the visibility labels for each  $i$-th training image, $\{\vv_i\}_{i=1}^N$, estimated by the the MNN. The initial shape is progressively refined in the cascade in $S$ stages by extracting shape indexed features on the heatmaps $\{\phi(\vh_i, \vv_i, \vx^{s-1}_i)\}_{i=1}^N$ following a coarse-to-fine procedure like~\cite{Valle19b}, where $\vx^{s-1}_i$ represents the shape of the $i$-th sample on the previous stage. The novelty of OR is that, the 2D displacements estimated by trees whose associated landmark is occluded are not added to the final estimation (see Fig.~\ref{fig:oert}).


\section{Experiments}
\label{sec:experiments}

To evaluate our approach we perform experiments using four in-the-wild landmark-related data bases and one head pose data set acquired in laboratory conditions.
COFW~\cite{Burgos13} focuses on occlusions. It provides 1345 faces annotated with the positions and the binary occlusion labels for 29 landmarks. On average 28\% of the landmarks are occluded. 
AFLW~\cite{Koestinger11} provides a collection of 25993 faces, with 21 facial landmarks annotated depending on their visibility. For our experiments we discard some images with reported annotation errors~\cite{Jin15}. We divide AFLW test subset into intervals of [0$^{\circ}$, 30$^{\circ}$], [30$^{\circ}$, 60$^{\circ}$] and [60$^{\circ}$, 90$^{\circ}$] according to head absolute yaw angle. 
AFLW2000-3D~\cite{Zhu17} consists of 2000 faces from AFLW semi-automatically re-annotated with 68 3D facial landmarks. We divide it into intervals of [0$^{\circ}$, 30$^{\circ}$], [30$^{\circ}$, 60$^{\circ}$] and [60$^{\circ}$, 90$^{\circ}$]. Each interval consists of 1306, 462 and 232 faces respectively. It has been typically used for testing head pose and facial landmark location algorithms using 300W-LP as train set~\cite{Zhu17}. This last data set provides 61225 synthesized face images from 300W~\cite{Sagonas16}, also re-annotated with 68 3D landmarks using the same algorithm.
The semi-automatic pipeline used to label 300W-LP and AFLW2000-3D has been criticised for not producing accurate annotations for extreme poses and occluded faces~\cite{Bulat17b}. For this reason we only use 300W-LP/AFLW2000-3D  for comparing with the state-of-the-art that follows this protocol.

Although it was not acquired in-the-wild, we also evaluate our model with
Biwi-Kinect~\cite{Fanelli13}. It contains 15677 images from 24 sequences of 20 subjects
acquired in a controlled environment with a Kinect sensor.
Since Biwi does not contain landmark annotations, we follow the protocol presented in~\cite{Ruiz18} using 300W-LP as train set.

\subsection{Evaluation Metrics}
We use the Mean Absolute Error (\textrm{MAE}) metric to quantify the head pose estimation error,
\begin{equation}
\textrm{MAE} = \frac{1}{N} \sum\limits_{i=1}^{N} \left( |\tilde{\vp}_i-\vp_i| \right),
\label{eq:mae}
\end{equation}
where N is the number of face images and $\tilde{\vp}_i$, $\vp_i$ represent the ground truth and predicted pose parameters respectively. 

We also use the Normalized Mean Error (\textrm{NME}) as a metric to measure the shape estimation error
\begin{equation}
        \textrm{NME} = \frac{100}{N} \sum\limits_{i=1}^{N} \Bigg( 
        \sum\limits_{l=1}^{L} \Big( \frac{\tilde{\vw}_i^l}{||\tilde{\vw}_i||_1}\cdot{}  \frac{||{\vx_i^l-\tilde{\vx}_i^l}||_2}{d_i} \Big) \Bigg),
    \label{eq:nme}
\end{equation}
where $\tilde{\vx}_i$ and $\vx_i$ are respectively the ground truth and estimated shape for the $i$-th training image and $d_i$ is a normalization value. We use different values of $d_i$: the distance between eye pupils (\emph{pupils}) and the bounding box height (\emph{height}). 

Finally, we report recall percentage at 80\% precision to compare landmarks visibility prediction with other published methods.

\subsection{Implementation Details}
\label{sec:implementation}

We train our models using Adam with an initial learning rate $\alpha=10^{-3}$, which is halved whenever the loss plateaus for 15 epochs.
We shuffle each training set and split it into 90\% train and 10\% validation.
%
We also augment our training data by applying to each sample the following random operations: in plane rotation between $\pm45^\circ$, scaling by $\pm15\%$, translation by $\pm5\%$ of the bounding box size, mirroring face image horizontally and colour change multiplying each HSV channel by a random value between $[0.5, 1.5]$. Additionally, we include synthetic rectangular occlusions to enforce the encoder-decoder to learn visibility.

When provided we crop faces using the data set bounding box annotations. In 300W-LP/AFLW2000-3D and Biwi we use respectively the rectangle enclosing the annotated landmarks and the thresholded depth image. These detections are enlarged by 30\% and resized to 256$\times$256 pixels.
In the landmark-annotated data sets we use POSIT~\cite{DeMenthon95} with a set of 2D (image) and 3D (face model) landmark correspondences to compute the head pose. We use as model the mean 3D face shape provided with AFLW~\cite{Koestinger11}. 

At runtime our implementation of MNN+OR processes test images on average at a rate of 12.8 FPS using a NVidia GeForce GTX 1080Ti (11GB) GPU and a dual Intel Xeon Silver 4114 CPU at 2.20GHz (2$\times{}$10 cores/40 threads, 128 GB), where the MNN takes 66 ms and the OR 12 ms per face using C++, Tensorflow and OpenCV libraries. 
We may also dispose of the MNN decoder and the OR regressor to build a very efficient head pose estimation module. The resulting model infers head pose using the GPU at a rate of 62.5 FPS.

\subsection{Ablation study}
\label{sec:subsection_ablation_multitask}

In this section, we analyze the contribution of each component in our framework in the final performance.

\subsubsection{Task location}

\begin{figure}
\centering
\includegraphics[width=0.49\textwidth]{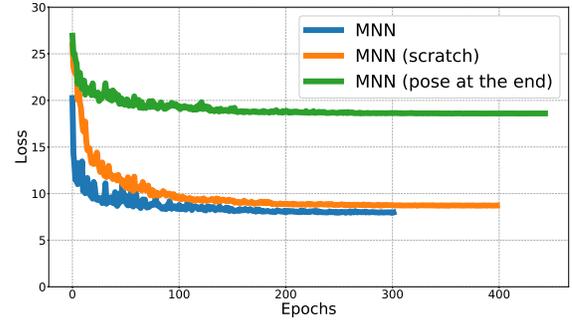}
\caption{Blue, orange and green colored learning curves compare the overall validation loss, $\mathcal{L}$, obtained with MNN by fine-tuning from landmarks, training from scratch, and locating the rigid pose losses at the end of the decoder respectively.}
\label{fig:learning_curves}
\end{figure}

In the first experiment we evaluate the importance of locating the head pose losses at the MNN bottleneck. To this end we adopt a MTL strategy pre-training the model with the landmark location task. The green and blue curves in Fig.~\ref{fig:learning_curves} show respectively the loss achieved when locating both rigid pose losses, $\mathcal{L}_{ED}$ and $\mathcal{L}_{AE}$, at the end of the decoder ($\mathcal{L}=18.6$) and at the end of the encoder ($\mathcal{L}=7.9$). 
In the second case we achieve a reduction of 57.5\% in the final loss. We infer that this gain is caused by two reasons. First, the superiority of the holistic features extracted from the embedding in the encoder-decoder bottleneck. Second, because head pose and landmark-related tasks do not co-operate when located in the same layer. Hard parameter sharing among these tasks decreases the final performance. 
From now on, we attach the rigid head estimation losses, $\mathcal{L}_{ED}$ and $\mathcal{L}_{AE}$, at the end of the encoder.

\subsubsection{Training strategy}

For these experiments, we incorporate two 2D landmark-based in-the-wild data sets. 300W~\cite{Sagonas16} provides 68 manually annotated facial landmarks. We followed the most established approach and divide the annotations into 3148 training and 689 testing images (public competition). Thereafter, we also perform experiments on the 300W private benchmark, using previous 3837 images for training and 600 newly updated images as testing set. WFLW~\cite{Wu18a} consists of 7500 extremely challenging training and 2500 testing images divided into six subgroups, pose, expression, illumination, make-up, occlusion and blur, with 98 fully manual annotated landmarks. Since these data sets do not provide any head pose label, we compute it using POSIT~\cite{DeMenthon95} with AFLW~\cite{Koestinger11} mean 3D face shape.

Here we evaluate our model under different training strategies. In the simplest case we follow a single task approach and minimize  $\mathcal{L}_{ED}(\vp^{ED})$ in Eq.~\eqref{eq:euclidean_distance_loss_min} (Pose row in Table~\ref{table:multitask_improvement}).
We also consider several symmetric and asymmetric MTL schemes. In the symmetric case we train our model from scratch with all three tasks, minimizing $\mathcal{L}(\vp^{ED}, \vp^{AE}, \vh, \vv)$ in \eqref{eq:global_loss} (Sym row in Table~\ref{table:multitask_improvement}, orange stroke in Fig.~\ref{fig:learning_curves}). 
We also look at an asymmetric MTL scheme in which we pre-train  with the image alignment task, optimizing $\mathcal{L}_{CE}(\vh)$ in Eq.~\eqref{eq:loss_crossentropy}. Once this training converges, we include the head pose and visibility tasks and optimize $\mathcal{L}(\vp^{ED}, \vp^{AE}, \vh, \vv)$ (Pre+Sym row in Table~\ref{table:multitask_improvement}, blue stroke in Fig.~\ref{fig:learning_curves}).
Finally, in the most asymmetric MTL situation, we pre-train with the image alignment task, optimizing $\mathcal{L}_{CE}(\vh)$. Upon convergence, we then only optimize the head pose task, $\mathcal{L}_{ED}(\vp^{ED})$ (Pre+Pose row in Table~\ref{table:multitask_improvement}). 

The orange and blue curves in Fig.~\ref{fig:learning_curves} respectively display the difference between using the symmetric MTL training scheme ($\mathcal{L}=8.7$) against the asymmetric MTL that pre-trains with the landmarks task followed by a symmetric MTL with all three tasks ($\mathcal{L}=7.9$). In our problem pre-training regularizes the learning process and achieves a 9\% reduction in the final loss, $\mathcal{L}$. 

Further, in Table~\ref{table:multitask_improvement} we show head pose estimation results for different landmark-based data sets and training strategies. On average we achieve the largest improvement in mean \textrm{MAE} when changing from single task learning (first row) to MTL (three bottom rows). In the worst case, when moving from single task to the symmetric MTL case, we achieve a 7.5\% reduction in mean \textrm{MAE}. 
The asymmetric approaches, that involve a pre-training step with the landmark face alignment task, achieve the best results, with a reduction of 11.9\% in the average mean \textrm{MAE} with respect to the single task approach. 
%
There is no difference whether after pre-training we refine the model only for the pose task or for all three tasks.
Hence, the second model will be the selected configuration and training strategy in our experiments.


\begin{table}[htbp!]
\begin{center}
\setlength\tabcolsep{1.7pt}
\begin{tabular}{l|l|c|c|c|c|c||c}
\hline
\multicolumn{2}{c|}{Method} & 300W pub & 300W priv & COFW & AFLW & WFLW & Avg\\
\hline
Single task & Pose & 1.91 & 2.22 & 2.67 & 3.43 & 2.46 & 2.54\\ 
\hline
\multirow{3}{*}{Multi-task} & Sym & 1.76 & 1.97 & 2.57 & 3.35 & 2.10 & 2.35\\
& Pre+Sym & 1.59 & 1.96 & 2.36 & 3.22 & 2.08 & 2.24\\
& Pre+Pose & 1.56 & 1.96 & 2.34 & 3.23 & 2.11 & 2.24\\
\hline
\end{tabular}
\end{center}
\caption{Head pose mean MAEs for different training strategies. First row (Pose) single task encoder. Second row (Sym) symmetric MTL for all three tasks. Third row (Pre+Sym) MTL learning scheme pre-training with face landmarks. Fourth row (Pre+Pose) asymmetric MTL scheme pre-training with landmarks fine-tuned with pose.}
\label{table:multitask_improvement}
\end{table}

We also evaluate the importance of the MTL scheme for visibility estimation using the COFW data set. Fist, we train MNN only for the visibility task. In this case, we achieve a recall of 21.75\% at 80\% precision for occlusion detection. This is a poor result, far worse than most published results (see Table~\ref{table:visibility_cofw}),  possibly caused by the small size of the training data set. However, using our selected MTL strategy, we get a recall of 72.12\%, a large improvement in recall at the typical 80\% precision point. So, with a small training data set, such as COFW, the combination of multiple related tasks within a MTL scheme boosts the final performance. 

\subsubsection{Occlusion-aware regressor}

Here we evaluate the contribution of the OR stage to the performance of the landmark location task. We report the \textrm{NME} of, 1) MNN alone, locating each landmark by the maximum response on its heatmap; 2) the full MNN+OR framework. We show in Tables~\ref{table:multitask_cofw},~\ref{table:multitask_aflw}  and~\ref{table:multitask_300wlp} the performances in COFW, AFLW and AFLW2000-3D data sets.
These results prove the importance of the OR stage to regularize the MNN landmark predictions. Model MNN+OR reduces the NME of model MNN in COFW by 10.8\%, in AFLW by 3\% and in AFLW2000-3D by 6.9\%. The improvement grows proportionally with the presence of self-occluded parts (\ie AFLW2000-3D) and non-visible landmarks (\ie COFW) in the data sets.

\subsection{Comparison with the state-of-the-art}
In this section we evaluate our model in the most challenging benchmarks for all three tasks.


\subsubsection{Head pose}

In Table~\ref{table:multitask_headpose_results} we compare our head pose estimation proposal with the best published results in the literature. We train two MNN models. For AFLW there is no standard protocol to determine the training and testing partitions. We use the benchmark proposed in~\cite{Amador17}. For testing in AFLW2000-3D and Biwi we train our model with 300W-LP, like~\cite{Ruiz18,Liu19b,Hsu18,Yang19,Zhang20}. However, we use the pose estimated from the correct 300W-LP landmarks from~\cite{Bulat17b}.

\begin{table*}[htbp!]
\begin{center}
\setlength\tabcolsep{5.0pt}
\begin{tabular}{l|ccc|c|ccc|c|ccc|c|ccc|c}
\hline
\multirow{2}{*}{Method} & \multicolumn{4}{c|}{AFLW} & \multicolumn{4}{c|}{AFLW2000-3D} & \multicolumn{4}{c|}{AFLW2000-3D-POSIT} & \multicolumn{4}{c}{Biwi}\\
 & yaw & pitch & roll & mean & yaw & pitch & roll & mean & yaw & pitch & roll & mean & yaw & pitch & roll & mean\\
 \hline
FAb-Net~\cite{Wiles18b} & 10.70 & 7.13 & 5.14 & 7.65 & - & - & - & - & - & - & - & - & - & - & - & -\\
Kepler~\cite{Kumar18b} & 6.45 & 5.85 & 8.75 & 7.01 & - & - & - & - & - & - & - & - & 8.08 & 17.2 & 16.1 & 13.8\\
Hyperface~\cite{Ranjan19} & 7.61 & 6.13 & 3.92 & 5.88 & - & - & - & - & - & - & - & - & - & - & - & -\\
HopeNet~\cite{Ruiz18} & 6.26 & 5.89 & 3.82 & 5.32 & 6.47 & 6.55 & 5.43 & 6.15 & - & - & - & - & 4.81 & 6.60 & 3.26 & 4.89\\
GLDL~\cite{Liu19b} & 6.00 & 5.31 & 3.75 & 5.02 & \textbf{3.02} & 5.06 & 3.68 & 3.92 & - & - & - & - & 4.12 & 5.61 & 3.14 & 4.29\\
HF-ResNet\cite{Ranjan19} & 6.24 & 5.33 & 3.29 & 4.95 & - & - & - & - & - & - & - & - & - & - & - & -\\
CCR~\cite{Zhang18c} & 5.22 & 5.85 & 2.51 & 4.52 & - & - & - & - & - & - & - & - & - & - & - & -\\
Amador \etal~\cite{Amador17} & 5.59 & 4.79 & 2.83 & 4.40 & - & - & - & - & - & - & - & - & - & - & - & -\\
FSA-Caps-Fusion~\cite{Yang19} & - & - & - & - & 4.50 & 6.08 & 4.64 & 5.07 & - & - & - & - & 4.27 & 4.96 & 2.76 & 4.00\\
QuatNet~\cite{Hsu18} & \textbf{3.93} & 4.31 & 2.59 & 3.61 & 3.97 & 5.61 & 3.92 & 4.50 & - & - & - & - & 4.01 & 5.49 & 2.93 & 4.14\\
FDN~\cite{Zhang20} & - & - & - & - & 3.78 & 5.61 & 3.88 & 4.42 & - & - & - & - & 4.52 & 4.70 & 2.56 & 3.93\\
\hline
MNN & 4.16 & \textbf{3.07} & \textbf{2.43} & \textbf{3.22} & 3.34 & \textbf{4.69} & \textbf{3.48} & \textbf{3.83} & \textbf{2.15} & \textbf{1.40} & \textbf{1.58} & \textbf{1.71} & \textbf{3.98} & \textbf{4.61} & \textbf{2.39} & \textbf{3.66}\\
\hline
\end{tabular}
\end{center}
\caption{Head pose MAEs for AFLW, AFLW2000-3D and Biwi. AFLW200-3D-POSIT is the outcome of re-annotating AFLW2000-3D with the corrected landmarks annotations from~\cite{Bulat17b}.}
\label{table:multitask_headpose_results}
\end{table*}

We outperform the state-of-the-art in AFLW (3.22 \textrm{MAE}), which represents an 11\% mean \textrm{MAE} reduction over QuatNet~\cite{Hsu18}, the best reported result in the literature. 
Moreover, in our MTL strategy we only use AFLW annotations, whereas QuatNet and many competing approaches use additional training data~\cite{Amador17,Ruiz18,Hsu18,Liu19b,Wiles18b}. 
%
In the experiment evaluated on AFLW2000-3D and Biwi our approach establishes two new top results, again, with no extra training data. While in Biwi we reduce in 6.9\%  FDN's~\cite{Zhang20} MAE, in AFLW2000-3D our result only improves by 2.3\% GLDL's~\cite{Liu19b}. This is caused by the inaccurate AFLW2000-3D annotations in extreme head poses~\cite{Bulat17b}. While our approach was trained with poses estimated from the corrected landmarks, our competitors were trained on the original 300W-LP annotations, poisoned with the same errors. We re-annotated AFLW2000-3D with the poses estimated from the correct landmarks. We denote this data set AFLW2000-3D-POSIT. In this case the mean MAE of our approach goes down to 1.71.

The results in Table~\ref{table:multitask_headpose_results} must be considered with caution. 
First, it seems obvious that landmark detection and head pose estimation tasks are clearly more connected if the pose is calculated from the landmarks. Moreover, the MAEs of AFLW and AFLW2000-3D-POSIT have a negative (optimist) bias. This is because in them the head pose in the train and test sets is computed with the same semi-automatic estimation procedure. The same argument applies to all our competitors in AFLW2000-3D. However, our results would be positively (pessimistically) biased in this data set, since some of its annotations are not correct. Hence, our unbiased MAE for AFLW2000-3D would be between the 3.83 and 1.71 bounds. In contrast, the experiment with Biwi involves different train/test data sets and annotation procedures. Hence, it provides the most accurate MAE estimations. Although, in a data set taken in laboratory conditions.

\subsubsection{Facial landmarks visibility}

In Table~\ref{table:visibility_cofw} we compare the landmarks visibility estimation that we obtain with MNN, against the best published results in the literature. To evaluate this task we use COFW~\cite{Burgos13}, as far as we known, the only data set with annotated occlusions.

\begin{table}[htbp!]
\begin{center}
\begin{tabular}{l|c}
\hline
\multirow{2}{*}{Method} & Full\\
 & occlusion\\
\hline
RCPR~\cite{Burgos13} & 40\\
Wu \etal~\cite{Wu17} & 44.43\\
Wu \etal~\cite{Wu15} & 49.11\\
ECT~\cite{Zhang18a} & 63.4\\
3DDE~\cite{Valle19b} & 63.89\\
\hline
MNN & \textbf{72.12}\\
\hline
\end{tabular}
\end{center}
\caption{Recall of landmarks visibility estimation methods at 80\% precision using COFW.}
\label{table:visibility_cofw}
\end{table} 

With our approach, we get a 72.12\% recall at 80\% precision, a new state-of-the-art for this data set. 
The notable improvement with respect to the closest competitor, 3DDE~\cite{Valle19b}, is caused by two key differences. First, the MTL strategy boost the performance of the landmark visibility task. Second, in 3DDE the visibility is estimated by the landmark ERT regressor, like in~\cite{Burgos13,Valle18}, whereas in the proposed approach the visibility is estimated in the MNN model.
This result proves again the relevance of our MTL approch. 

\subsubsection{Facial landmark location}

We compare the MNN+OR framework with the state-of-the-art in face landmark regression. To this end we use results reported for 2D and 3D face alignment data sets. We use COFW and AFLW to provide a reference comparison with data sets  involving 2D landmarks and AFLW2000-3D for 3D landmarks.

We analyze in Table~\ref{table:multitask_cofw} the MNN+OR landmark location performance in COFW, the common benchmark to evaluate occlusions. Here, we achieve a performance comparable to the best reported result for this data set, CHR2C~\cite{Valle19a}, based on two stacked U-Net-like models.

\begin{table}[htbp!]
\begin{center}
\begin{tabular}{l|c}
\hline
\multirow{2}{*}{Method} & Full\\
 & pupils\\
\hline
RCPR~\cite{Burgos13} & 8.50\\
TCDCN~\cite{Zhang16} & 8.05\\
Wu \etal~\cite{Wu17} & 6.40\\
Wu \etal~\cite{Wu15} & 5.93\\
ECT~\cite{Zhang18a} & 5.98\\
PCD-CNN~\cite{Kumar18a} & 5.77\\
SHN~\cite{Yang17} & 5.6\\
Wing~\cite{Feng18a} & 5.44\\
ODN~\cite{Zhu19} & 5.30\\
3DDE~\cite{Valle19b} & 5.11\\
CHR2C~\cite{Valle19a} & 5.09\\
\hline
MNN & 5.65\\
MNN+OR & \textbf{5.04}\\
\hline
\end{tabular}
\end{center}
\caption{Face alignment NME using COFW.}
\label{table:multitask_cofw}
\end{table}

In Table~\ref{table:multitask_aflw} we compare MNN+OR with previous literature using AFLW images. This is a challenging database due to the large number of faces with extreme poses and occluded landmarks, which are not annotated. 
In this case, again, we achieve a performance comparable to the best reported result in the literature.

\begin{table}[htbp!]
\begin{center}
\begin{tabular}{l|c|c|c|c}
\hline
\multirow{2}{*}{Method} & [0$^{\circ}$, 30$^{\circ}$] & [30$^{\circ}$, 60$^{\circ}$] & [60$^{\circ}$, 90$^{\circ}$] & Full\\
 & height & height & height & height\\
\hline
CCR~\cite{Zhang18c} & - & - & - & 5.72\\
Hyperface~\cite{Ranjan19} & 3.93 & 4.14 & 4.71 & 4.26\\
Kepler~\cite{Kumar18b} & - & - & - & 2.98\\
AIO~\cite{Ranjan17} & 2.84 & 2.94 & 3.09 & 2.96\\
HF-ResNet~\cite{Ranjan19} & 2.71 & 2.88 & 3.19 & 2.93\\ 
Binary-CNN~\cite{Bulat17a} & 2.77 & 2.60 & 2.64 & 2.85\\
PCD-CNN~\cite{Kumar18a} & 2.33 & 2.60 & 2.64 & 2.49\\
3DDE~\cite{Valle19b} & 2.10 & 2.00 & 2.04 & 2.06\\
CHR2C~\cite{Valle19a} & 2.07 & \textbf{1.86} & \textbf{1.81} & 1.98\\
\hline
MNN & 2.12 & 1.90 & 1.89 & 2.03\\
MNN+OR & \textbf{2.05} & \textbf{1.86} & 1.85 & \textbf{1.97}\\
\hline
\end{tabular}
\end{center}
\caption{Face alignment NME using AFLW.}
\label{table:multitask_aflw}
\end{table}


Finally, in Table~\ref{table:multitask_300wlp}, we also evaluate our model using a 3D data set. To this end we train our model with 300W-LP and test in AFLW2000-3D.
In this case, we achieve 2.58 \textrm{NME} in the \textit{Full} set. This result sets the new state-of-the-art for this data set, with a 16.2\% reduction in \textrm{NME} with respect to the best published result in the literature, MHM~\cite{Deng18} (3.08 \textrm{NME}), based on a two-stage cascade of heatmap regressors. Even without the final OR regressor, the MNN model alone already improves in 10\% the previous best result. 
Our two-stage hybrid strategy is specially effective in 3D face alignment, where the OR stage is initialized using the extremely accurate head pose estimated by the MNN
(see Fig.~\ref{fig:oert}).

\begin{table}[htbp!]
\begin{center}
\begin{tabular}{l|c|c|c|c}
\hline
\multirow{2}{*}{Method} & [0$^{\circ}$, 30$^{\circ}$] & [30$^{\circ}$, 60$^{\circ}$] & [60$^{\circ}$, 90$^{\circ}$] & Full\\
 & height & height & height & height\\
\hline
RCPR~\cite{Burgos13} & 4.26 & 5.96 & 13.18 & 7.80\\
3DSTN~\cite{Bhagavatula17} & 3.15 & 4.33 & 5.98 & 4.49\\
3DDFA~\cite{Zhu17} & 2.84 & 3.57 & 4.96 & 3.79\\
PRN~\cite{Feng18b} & 2.75 & 3.51 & 4.61 & 3.62\\
Binary-CNN~\cite{Bulat17a} & 2.47 & 3.01 & 4.31 & 3.26\\
MHM~\cite{Deng18} & \textbf{2.36} & 2.80 & 4.08 & 3.08\\
\hline
MNN & 2.71 & 2.53 & 3.48 & 2.77\\
MNN+OR & 2.54 & \textbf{2.24} & \textbf{3.34} & \textbf{2.58}\\
\hline
\end{tabular}
\end{center}
\caption{Face alignment NME using AFLW2000-3D.}
\label{table:multitask_300wlp}
\end{table}


\section{Conclusions}
\label{sec:conclusions}
In this paper we have presented a supervised multi-task approach to head pose, facial landmark location and visibility estimation. It is based on a heatmap encoder-decoder CNN, MNN, followed by an ensemble of regression trees to estimate the landmark co-ordinates. Rather than using head pose as a by-product or auxiliary task for landmark estimation, in our approach landmark-related tasks are used to boost head pose estimation. However, they are only required at training time. During testing we may dispose of the decoder and landmark regression modules to produce an extremely efficient head pose regressor with the best reported accuracy in the literature. In our head pose estimation experiments with landmark-based data sets we improve the best reported result in AFLW, QuatNet~\cite{Hsu18}, and GLDL~\cite{Liu19b}, the top performer in AFLW2000-3D. We also establish a new state-of-the-art in Biwi, a data set acquired in laboratory conditions and accurately annotated from depth images.  

The MNN model and the MTL training strategy are fundamental to achieve top performance with our framework. In the ablation analysis we show that we get the largest improvement when switching  from single task to a multi-task approach. We can further improve the performance if we adopt an asymmetric MTL scheme and pre-train the MNN with the face landmark estimation task. This confirms previous results showing that pre-training a model with a hard problem significantly improves the performance of other related tasks~\cite{Tran19}. Also, our ablation shows that hard parameter sharing between head pose and face landmark estimation is detrimental of the final performance. This also confirms that multi-task and transfer (pre-training) relationships are different~\cite{Standley19}. To address this issue and to provide each task with the appearance information aggregated at the best level of abstraction, we hook up the head pose loss to the encoder end, whereas the losses of spatially related tasks, such as landmark location and visibility, are attached to the decoder end. 

Our model also reaches top performance in the two landmark related tasks. In visibility estimation it achieves 72.12\% recall at 80\% precision in COFW. A 13\% improvement over the previous reported state-of-the-art, 3DDE~\cite{Valle19b}. We also compute the location of face landmarks using a novel occlusion-aware regressor (OR), that estimates face deformation from the heatmaps of visible landmarks. The full MNN+OR achieves results comparable to  the state-of-the-art, 3DDE and CHR2C~\cite{Valle19a,Valle19b}, when evaluated in AFLW and COFW. In AFLW2000-3D, where self-occlusions play a key role, it sets a reduction of 16\% over the previous state-of-the-art, MHM~\cite{Deng18}.

A fundamental problem to build a head pose estimation algorithm is the lack of training data. We propose a MTL strategy that takes advantage of the data bases available with manually labelled face landmarks. Pre-training with large object or face recognition data sets are alternative popular means to address this issue. We have proved that in the context of head pose estimation our proposal beats this strategy. This is due to the better co-operation between head pose and face landmark tasks. To further increase the robustness and accuracy of head pose estimation, our approach may be combined with self-supervised training~\cite{Wiles18b,Zhang18e,Thewlis19,Jeon19} and the use of synthetically generated data sets~\cite{Kuhnke19}.

It is difficult to establish an state-of-the-art for head pose estimation in-the-wild, due to the lack of accurately annotated data sets. 
Present in-the-wild head pose evaluation methodologies are based on landmark data bases, such as AFLW and 300W-LP/AFLW2000-3D. The semi-automatic pipeline used to process them introduces errors in the train and test set annotations that bias the evaluation. Using re-annotated 300W-LP/AFLW2000-3D head poses we were able to upper and lower bound the performance of our approach in this situation.
Biwi's evaluation methodology, based on train/test data sets acquired with different technologies and annotation algorithms, provides a more realistic performance estimation in laboratory conditions.

The proposed approach not only provides a satisfactory prediction accuracy but also a good computational efficiency. Instead of evaluating three different models, one for each task, we use a single encoder-decoder CNN, with an extremely efficient ERT, to simultaneously solve all three tasks at a rate of 12.8 FPS. However, if we are only interested in estimating head pose as a preliminary face processing step~\cite{Chang17}, our encoder-only model achieves 62.5 FPS.

\ifCLASSOPTIONcompsoc
  \section*{Acknowledgments}
\else
  \section*{Acknowledgment}
\fi
The authors acknowledge funding from the Spanish Ministry of Economy and Competitiveness, project TIN2016-75982-C2-2-R.
Jos{\'e} M. Buenaposada was also partially funded by the \emph{Comunidad de Madrid} project RoboCity2030-DIH-CM (S2018/NMT-4331). They also thank the anonymous reviewers for their comments and Felix Kuhnke for his help in interpreting Biwi  annotations.

%
%
\bibliographystyle{IEEEtran}
\bibliography{faces}

%








\end{document}